\documentclass[11pt,a4paper]{article}
\usepackage[hyperref]{emnlp2020}
\usepackage{times}
\usepackage{latexsym}

\usepackage{microtype}

\usepackage{booktabs}       
\usepackage{amsfonts}       
\usepackage{nicefrac}       
\usepackage{microtype}      
\usepackage{lipsum}
\usepackage{graphicx}
\usepackage{todonotes}
\usepackage{xspace}
\usepackage[hang,flushmargin]{footmisc}
\newcommand{\sig}{* }
\usepackage{enumitem}

\newcommand{\sys}{\textsc{Sledge-Z}\xspace}
\newcommand{\edit}[1]{#1}

\aclfinalcopy 

\title{SLEDGE-Z: A Zero-Shot Baseline for COVID-19 Literature Search}

\author{Sean MacAvaney\thanks{\hspace{0.6em}\edit{This work was done while at an internship at the Allen Institute for AI.}}\ \ $^\dag$\ \hspace{1.7em}
Arman Cohan$^\ddag$ \hspace{1.7em}
{\bf Nazli Goharian$^\dag$} \hspace{1.4em}
 \vspace{6pt}\\
  $^\dag$Information Retrieval Lab,
  Georgetown University, Washington DC  \vspace{2pt} \\
  $^\ddag$Allen Institute for AI, Seattle WA \vspace{2pt}\\
  {\tt \{sean,nazli\}@ir.cs.georgetown.edu, armanc@allenai.org}
}

\date{}

\begin{document}
\maketitle
\begin{abstract}
With worldwide concerns surrounding the Severe Acute Respiratory Syndrome Coronavirus 2 (SARS-CoV-2), there is a rapidly growing body of scientific literature on the virus. Clinicians, researchers, and policy-makers need to be able to search these articles effectively. In this work, we present a zero-shot ranking algorithm that adapts to COVID-related scientific literature. Our approach filters training data from another collection down to medical-related queries, uses a neural re-ranking model pre-trained on scientific text (SciBERT), and filters the target document collection. This approach ranks top among zero-shot methods on the TREC COVID Round 1 leaderboard, and exhibits a P@5 of 0.80 and an nDCG@10 of 0.68 when evaluated on both Round 1 and 2 judgments. \edit{Despite not relying on TREC-COVID data, our method outperforms models that do.} As one of the first search methods to thoroughly evaluate COVID-19 search, we hope that this serves as a strong baseline and helps in the global crisis. 
\end{abstract}


\section{Introduction}

The emergence of the Severe Acute Respiratory Syndrome Coronavirus 2 (SARS-CoV-2) prompted a worldwide research response. In the first 120 days of 2020, researchers published over 10,000 articles related to SARS-CoV-2 or COVID-19. Together with articles about similar viruses researched before 2020, the body of research approaches 60,000 articles.  Such a large body of research results in a considerable burden for those seeking information about various facets of the virus, including researchers, clinicians, and policy-makers.

\begin{figure}[t]
  \centering
  \includegraphics[width=\linewidth]{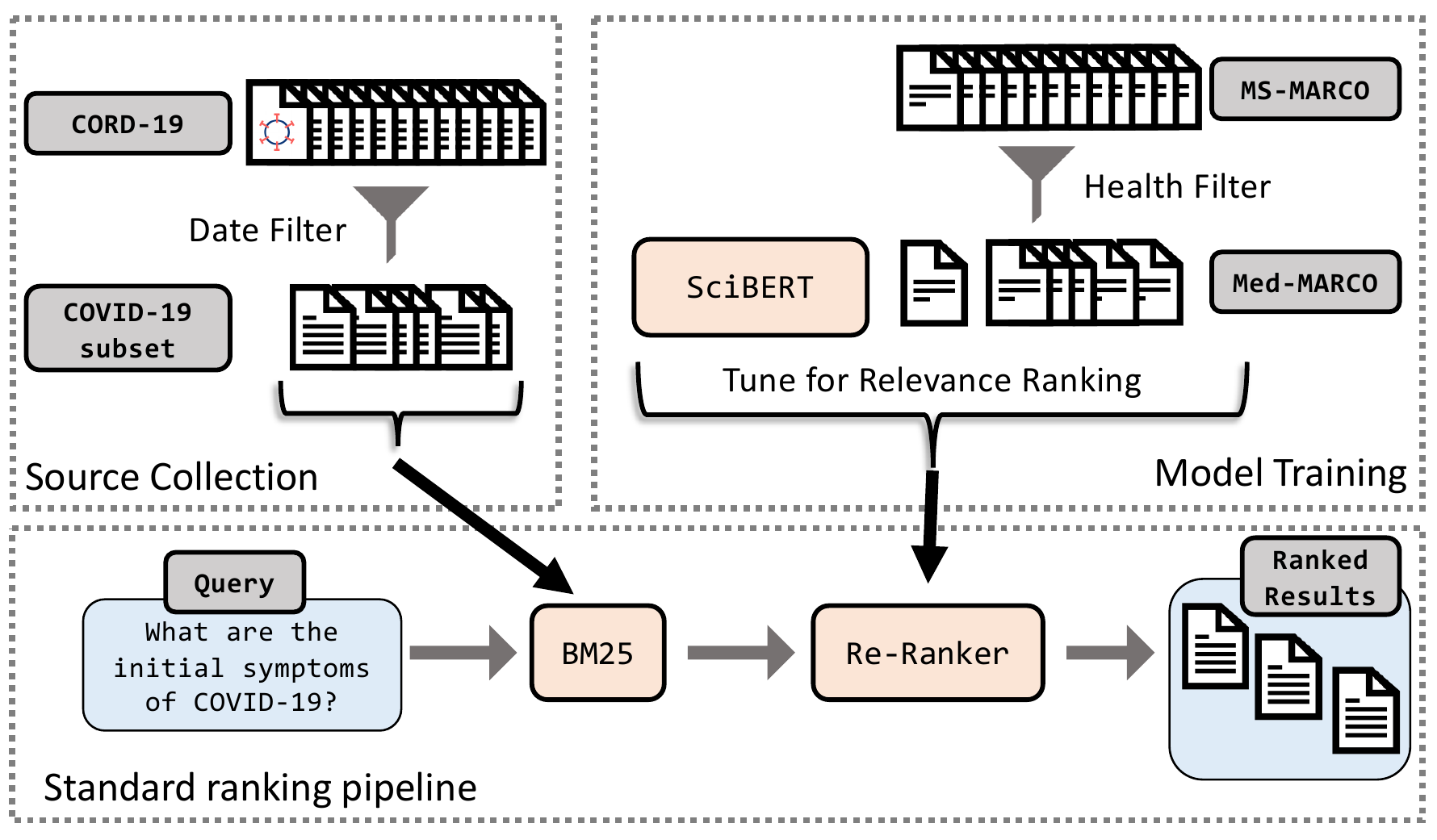}
  \caption{Overview of \sys.}
  \label{fig:sledge-fig}
\end{figure}

To help improve COVID-19 search, we introduce \sys: a simple yet effective zero-shot baseline for coronavirus \underline{S}cientific know\underline{LEDGE} search. \sys adapts the successful BERT-based \cite{Devlin2019BERTPO} re-ranking model (Vanilla BERT, \citet{MacAvaney2019CEDRCE}) for COVID-19 search with three simple techniques. First, we propose a training data filtering technique to help the ranking model learn relevance signals typical in medical text. The training data we use comes entirely from another dataset (MS-MARCO, \citet{Campos2016MSMA}), resulting in our model being zero-shot. Since MS-MARCO is a large collection of real user queries (over 800,000), it allows us to filter aggressively and still have adequate training data. Second, we replace the general contextualized language model BERT with one pre-trained on scientific literature (SciBERT, \citet{Beltagy2019SciBERTPC}). This pre-training prepares the model for the type of language typically seen in scientific articles. Since the document collection (CORD-19, \citet{Wang2020CORD19TC}) contains articles about prior viruses, we filter out articles published before 2020 to eliminate less pertinent articles. An overview of this process is shown in Figure~\ref{fig:sledge-fig}.

We show that each of the techniques mentioned above positively impacts the ranking effectiveness of \sys through an ablation analysis. Our zero-shot approach performs comparably to (or outperforms) top-scoring submissions to the TREC-COVID document ranking shared task~\cite{trec-covid}, a new testbed for evaluating of search methods for COVID-19. \sys tops the Round~1 leaderboard in the zero-shot setting, which is important in low-resource situations.
Overall, our method establishes a strong performance for COVID-19 literature search. By releasing our models and code, we hope that it can help in the current global COVID-19 crisis.\footnote{Code and models available at: \url{https://github.com/Georgetown-IR-Lab/covid-neural-ir}.}

\section{Related Work}

Ad-hoc document retrieval (of both scientific articles and general domain documents) has been long-studied~\cite{Lalmas2007INEX2,hersh2009trec,Lin2008IsSF,medlar2016pulp,sorkhei2017exploring,huang2019holes,Hofsttter2020InterpretableT,nogueira2020document}. Most recent work for scientific literature retrieval has focused on tasks such as collaborative filtering~\cite{Chen2018ResearchPR}, citation recommendation~\cite{Nogueira2020EvaluatingPT}, and clinical decision support~\cite{Soldaini2017LearningTR}.

Pre-trained neural language models (such as BERT~\cite{Devlin2019BERTPO}) have recently shown to be effective when fine-tuned for ad-hoc ranking \cite{Nogueira2019PassageRW,Dai2019DeeperTU,MacAvaney2019CEDRCE}.
These models also facilitate relevance signal transfer; \citet{Yilmaz2019CrossDomainMO} demonstrate that the relevance signals learned from BERT can transfer across collections (reducing the chance of overfitting a particular collection).
Here, we use relevance signal transfer from an open-domain question answering dataset to the collection of COVID-19 scientific literature.

Others have investigated COVID-19 document ranking. \citet{zhang2020rapidly} chronicled their efforts to build a search engine for COVID-19 articles, using a variety of available ranking techniques, such as T5~\cite{Raffel2019ExploringTL}. In this work, we find that our approach outperforms this system in terms of ranking effectiveness. \edit{Contemporaneously with our work, \citet{Das2020InformationRA} demonstrate how document clustering and summarization can be effective for COVID-19 retrieval. This paper extends our shared task submissions in Round 1~\cite{macavaney:arxiv2020-sledge}. We note that the TREC COVID task proceeded for a total of 5 rounds, with various techniques emerging, such as passage aggregation~\cite{Li2020PARADEPR,Nguyen2020Searching}, and ensemble methods~\cite{Bendersky2020RBF102}.}

\section{\sys: Zero-Shot COVID-19 Search}

To build a ranking model for COVID search, we modify the standard zero-shot Vanilla BERT document re-ranking pipeline~\cite{Yilmaz2019CrossDomainMO,MacAvaney2019CEDRCE}. We find that while these modifications are simple, they are effective for maximizing ranking performance. We note that this process neither requires COVID relevance training data nor involves a priori inspection of the queries and their characteristics. Thus, we consider our method zero-shot.

To train in a zero-shot setting, we employ a large dataset of general-domain natural language question and answer paragraphs: MS-MARCO~\cite{Campos2016MSMA}. However, na\"ive domain transfer is not optimal since most questions in the dataset are not medical-related, causing a domain mismatch between the training and evaluation data. To overcome this challenge, we apply a heuristic to filter the collection to only medical-related questions. The filter removes questions that do not contain terms appearing in the MedSyn~\cite{Yates2013ADRTraceDE}, a lexicon of layperson and expert terminology for various medical conditions. \edit{We manually remove several common terms from the lexicon that commonly introduce queries that are not medical-related. For example, MedSyn includes the term \textit{gas} (referring to the medical concept of flatulence in North American English), commonly also refers to gasoline or natural gas. See Appendix~\ref{sec:sup:medsyn-excl} for a complete list of excluded MedSyn terms. Note that we made these decisions without considering COVID-19 specifically---only a broad relation to the medical domain.} MS-MARCO originally consists of 809K questions. After filtering, 79K of the original questions remain (9.7\%). We refer to this subset of MS-MARCO as Med-MARCO. From a random sample of 100 queries from Med-MARCO, 78 were judged by the authors as medical-related, suggesting the filter has reasonable precision. \edit{Examples questions from this process include \textit{causes of peritoneal cancer prognosis} and \textit{what is squalene anthrax sleep apnea}.} We make a list of the query IDs corresponding to Med-MARCO available,\footnote{\url{https://github.com/Georgetown-IR-Lab/covid-neural-ir/blob/master/med-msmarco-train.txt}} as well as additional examples of filtered queries (see Appendix~\ref{sec:sup:filter-examples}).

Second, we replace the general-language BERT model with a variant tuned on scientific literature (including medical literature). Specifically, we use SciBERT~\cite{Beltagy2019SciBERTPC}, which has an identical structure as BERT, but was trained on a multi-domain corpus of scientific publications. It also uses a WordPiece lexicon based on the training data, allowing the model to better account for subwords commonly found in scientific text. During model training, we employ the pairwise cross-entropy loss function from~\citet{Nogueira2019PassageRW}. Relevant and non-relevant documents are sampled in sequence from the official MS-MARCO training pair list (filtered down to Med-MARCO queries).

Third, we apply a filter to the document collection that removes any articles published before January 1, 2020. This filter aims to improve the retrieval system's precision by eliminating articles that may discuss other topics. The date was chosen because little was known about COVID-19 prior to 2020, and some documents do not include a full publication date (only a year), making this filter simple to apply. In real-life search engines, date filtering can often be applied at the discretion of the user.

\section{Experimental setup}\label{sec:exp}



\begin{table*}
\centering\small
\scalebox{0.95}{
\begin{tabular}{llrrrrrrr}
\toprule
& & \multicolumn{4}{c}{Including Unjudged} & \multicolumn{3}{c}{Judged Only} \\
\cmidrule(lr){3-6} \cmidrule(lr){7-9}
Model & Training & nDCG@10 & P@5 & P@5 (F) & J@10 & nDCG@10 & P@5 & P@5 (F) \\
\midrule
BM25	&	 - &	 \sig 0.368 &	 \sig 0.469 &	 \sig 0.331 &	 75\% &	 \sig 0.436 &	 \sig 0.520 &	 \sig 0.383 \\
 + BERT 	&	 MSM &	 \sig 0.547 &	 \sig 0.617 &	 \sig 0.480 &	 83\% &	 \sig 0.617 &	 \sig 0.703 &	 \sig 0.549 \\
 + BERT 	&	 MedM &	 0.625 &	 \sig 0.697 &	 \sig 0.571 &	 92\% &	 0.657 &	 \sig 0.737 &	 \sig 0.606 \\
 + SciBERT &	 MSM &	 0.667 &	 0.754 &	 0.611 &	 88\% &	 \bf 0.724 &	 \sig 0.789 &	 0.646 \\
 + SciBERT (\sys) &	 MedM &	 \bf 0.681 &	 \bf 0.800 &	 \bf 0.663 &	 90\% &	 0.719 &	 \bf 0.846 &	 \bf 0.697 \\
\midrule
 + ConvKNRM & MSM & 0.536 & 0.617 & 0.491 & 86\% & 0.580 & 0.645 & 0.508 \\
 + ConvKNRM & MedM & 0.565 & 0.668 & 0.525 & 86\% & 0.621 & 0.714 & 0.565 \\
 + CEDR-KNRM & MSM & 0.514 & 0.617 & 0.468 & 86\% & 0.524 & 0.628 & 0.474 \\
 + CEDR-KNRM & MedM & 0.619 & 0.714 & 0.560 & 89\% & 0.649 & 0.742 & 0.582 \\
 + Seq2seq T5 & MSM & 0.656 & 0.737 & 0.634 & 90\% & 0.685 & 0.765 & 0.651 \\
 + Seq2seq T5 & MedM & 0.626 & 0.714 & 0.594 & 86\% & 0.678 & 0.754 & 0.628 \\
Fusion1 & - & 0.519 & 0.640 & 0.457 & 94\% & 0.534 & 0.640 & 0.457 \\
Fusion2 & - & 0.601 & 0.737 & 0.565 & 96\% & 0.605 & 0.737 & 0.565 \\
\bottomrule
\end{tabular}
}
\caption{\edit{Ablation results and comparison of our approach and other zero-shot baselines on TREC-COVID Rounds 1 and 2. The top results are shown in bold. SciBERT with MedM (\sys) significantly outperforms values in the top (ablation) section marked with \sig ($p<0.05$, paired t-test, Bonferroni correction).}}
\label{tab:main}
\end{table*}

We now explore the ranking effectiveness of our approach. We evaluate the performance of \sys using Round 1 and 2. \edit{At the time of writing, the only training data available for the task was the Round 1 data.} of the TREC-COVID Information Retrieval Benchmark~\cite{trec-covid}.\footnote{Round 2 uses \textit{residual collection} evaluation, meaning that all documents judged in Round 1 are disregarded. Although this is an important setting for building up a dataset and allows for approaches like manual relevance feedback, we feel that this setting does not mimic an actual search engine, especially in the zero-shot setting. Thus, we evaluate on the concatenation of Round 1 and 2 settings and mark the systems that use Round 1 judgments for training or tuning of their system.} TREC-COVID uses the CORD-19 document collection~\cite{Wang2020CORD19TC} (2020-05-01 version, 59,943 articles), with a set of 35 topics related to COVID-19. These topics include natural questions such as: \textit{what is the origin of COVID-19} and \textit{how does the coronavirus respond to changes in the weather}. The top articles of participating systems in each round were judged by expert assessors, who rated each article as non-relevant (0), partially-relevant (1), or fully-relevant (2) to the topic. In total, 20,728 relevance judgments were collected (avg. 592 per topic), with 74\% non-relevant, 12\% partially relevant, and 14\% fully-relevant. These rates remained nearly constant between rounds 1 and 2.

We use normalized Discounted Cumulative Gain with a cutoff of 10 (nDCG@10), Precision at 5 of partially and fully-relevant documents (P@5), and Precision at 5 of only fully relevant documents (P@5 (F)). Both nDCG@10 and P@5 are official task metrics; we include the P@5 filtered to only fully-relevance documents because it exposed some interesting trends in our analysis. We also report the percentage of the top 10 documents for each query that have relevance judgments (J@10). In an additional evaluation, we measure the performance using only judged documents to ensure that unjudged documents do not impact our findings. We used \texttt{trec\_eval}\footnote{\url{https://github.com/usnistgov/trec\_eval}} for all metrics. These measures represent a precision-focused evaluation; since re-ranking methods like ours focus on improving precision, we leave recall-oriented evaluations to future work.

Our initial ranking is conducted using BM25 with default settings over the full document text\edit{ to adhere to the zero-shot setting}. Re-ranking is conducted over the abstracts only\edit{, avoiding the need to perform score aggregation (since BERT models are limited in the document length)}. We utilize only the natural-language question (ignoring the keyword query and extended narrative). We conduct an ablation that compares \sys to versions using BERT (instead of SciBERT), and the full MS-MARCO dataset (MSM) (rather than the Med-MARCO subset (MedM)). We compare with several baselines under the same evaluation settings. 

\begin{itemize}[leftmargin=*]
\setlength{\itemsep}{0pt}
\setlength{\parskip}{0pt}
\item[-] \textbf{BM25}: the initial BM25 ranking.
\item[-] \edit{\textbf{ConvKNRM}: The convolutional KNRM model~\cite{Dai2018ConvolutionalNN}, trained on MS-MARCO data.}
\item[-] \edit{\textbf{CEDR KNRM}: The KNRM model, augmented with contextualized embeddings~\cite{MacAvaney2019CEDRCE}, trained on MS-MARCO data. We use the \texttt{bert-base-uncased} model for the contextualized embeddings.}
\item[-] \edit{\textbf{Seq2seq T5}: The text-to-text-transformer (T5) model~\cite{Raffel2019ExploringTL}, tuned for ranking by predicting \textit{true} or \textit{false} as the next term in a sequence consisting of the query and document~\cite{Nogueira2020DocumentRW}.}
\item[-] \textbf{Fusion}: a reciprocal rank fusion method~\cite{Cormack2009ReciprocalRF} of BM25 over the abstract, full text, and individual paragraphs. Fusion1 uses a concatenation of the keywords and question, and Fusion2 uses the entity extraction technique from the Round 1 \texttt{udel} submission.\footnote{\url{https://github.com/castorini/anserini/blob/master/docs/experiments-covid.md}}
\end{itemize}

Our work utilizes a variety of existing open-source tools: OpenNIR~\cite{macavaney:wsdm2020-onir}, Anserini~\cite{Yang2017AnseriniET}, and the HuggingFace Transformers library~\cite{Wolf2019HuggingFacesTS}. We utilize a held-out subset of 200 queries from the MS-MARCO training set as a validation set for the sole purpose of picking the optimal training epoch. Model hyper-parameters were chosen from values in prior work and can be found in Appendix~\ref{sec:sup:params}\edit{, along with information about the hardware used}. The Vanilla BERT and SciBERT models take approximately 3 hours to train/validate, and inference on TREC-COVID takes approximately 15 minutes on modern GPUs. The BERT model has 157M parameters, and the SciBERT model has 158M parameters.

\section{Results}\label{sec:res}

Ranking effectiveness is presented in Table~\ref{tab:main}. We first compare the ablations of our approach (top section). We note that SciBERT significantly ($p<0.05$, paired t-test, Boneferroni correction) outperforms BM25 and BERT trained on MSM across all metrics. There is a less dramatic jump between BERT MSM and BERT MedM, demonstrating the importance of filtering the training data properly. This is echoed between SciBERT MSM and SciBERT MedM, though the difference is only significant for P@5 when only considering the judged documents. These results demonstrate the importance of both pre-training on appropriate data and fine-tuning using a proper subset of the larger data. While both yield improvements (that can be additive), the pre-training objective appears to be more impactful, based on the overall better scores of SciBERT.

\begin{table*}
\centering\small
\scalebox{0.95}{
\begin{tabular}{llrrrrrrr}
\toprule
& & \multicolumn{4}{c}{Including Unjudged} & \multicolumn{3}{c}{Judged Only} \\
\cmidrule(lr){3-6} \cmidrule(lr){7-9}
Model & Training & nDCG@10 & P@5 & P@5 (F) & J@10 & nDCG@10 & P@5 & P@5 (F) \\
\midrule
\sys (ours) &	 MedM &	 0.681 &	 0.800 &	 0.663 &	 90\% &	 0.719 &	 0.846 &	 \bf 0.697 \\
\texttt{covidex.t5}$^\dagger$ & MSM, MedM & 0.618 & 0.731 & 0.560 & 94\% & 0.643 & 0.731 & 0.560 \\
\multicolumn{2}{l}{\hspace{1em}with date filter} & 0.652 & 0.760 & 0.600 & 92\% & 0.680 & 0.777 & 0.611 \\
\texttt{SparseDenseSciBert}$^\dagger$ & MedM & 0.672 & 0.760 & 0.646 & 96\% & 0.692 & 0.760 & 0.646 \\
\multicolumn{2}{l}{\hspace{1em}with date filter} & \bf 0.699 & 0.805 &\bf 0.691 & 94\% & \bf 0.724 & 0.811 & 0.691 \\
\texttt{mpiid5\_run3}$^\dagger$ & MSM, Rnd1 & 0.684 & \bf 0.851 & 0.640 & 93\% & 0.719 & \bf 0.851 & 0.640 \\
\multicolumn{2}{l}{\hspace{1em}with date filter} & 0.679 & 0.834 & 0.657 & 90\% & 0.722 & 0.834 & 0.657 \\
\bottomrule
\end{tabular}
}
\caption{\edit{TREC COVID Round 1 and 2 comparison between \sys and other top official Round 2 submissions. We apply the date filter for a more complete comparison. Note that experimental differences exist between our system and these submissions, including the use of multiple topic fields and the utilization of Round 1 training data for training or tuning. The top result is marked in bold.}}
\label{tab:main2}
\end{table*}

\begin{table}
\centering\small
\scalebox{0.95}{
\begin{tabular}{lrrr}
\toprule
System & nDCG@10 & P@5 & P@5 (F) \\
\midrule
\sys (ours) & \bf 0.641 & 0.747 & \bf 0.633 \\
\texttt{sab20.1.meta.docs} & 0.608 & \bf 0.780 & 0.487 \\
\texttt{IRIT\_marked\_base} & 0.588 & 0.720 & 0.540 \\
\texttt{CSIROmedNIR} & 0.588 & 0.660 & 0.587 \\
\bottomrule
\end{tabular}
}
\caption{TREC-COVID Round 1 leaderboard (automatic systems). \sys outperforms the highest-scoring run in terms of nDCG@10 and P@5 (F).}
\label{tab:rnd1}
\end{table}

\edit{Compared to baseline systems (bottom section), we observe that \sys offers superior effectiveness. Specifically, we see that ConvKNRM, CEDR-KNRM, and Seq2seq T5 all improve upon the initial BM25 ranking. Training on Med-MARCO (rather than the full MS-MARCO) also improves each of the baselines, except, curiously, Seq2seq T5. This model may benefit from the larger amount of training data the full MS-MARCO dataset offers. Finally, both fusion methods outperform the base BM25 model. However, we note that these models utilize two fields available for each query: the keyword-based query and the full natural-language question text---a luxury not available in practical search environments. (Recall that \sys and the other baselines in Table~\ref{tab:main} only use the natural-language query.)}

\edit{We now compare our approach with the top-performing submissions to the TREC COVID shared task \edit{(many of which are not zero-shot methods)}. Full participating system descriptions are provided in Appendix~\ref{sec:sup:rnd1}. We note that these experimental settings for these runs differ from our main experiments. For instance, \texttt{mpiid5\_run3}~\cite{Li2020PARADEPR} and \texttt{SparseDenseSciBERT} use relevant information from Round 1 as training data, and \texttt{covidex.t5} uses combined keyword query and natural-language questions. Therefore, these performance metrics are not directly comparable to our zero-shot runs. Despite this, \sys still achieves competitive performance compared to these models. For instance, it consistently scores comparably or higher than \texttt{covidex.t5} (includes a more powerful language model, a more effective initial ranking model, and multiple topic fields) and \texttt{SparseDenseSciBert} (which uses neural approaches for the initial ranking stage). Our method even performs comparably to the \texttt{mpiid5.run3} model, which was trained directly on Round 1 judgments. Interestingly, we observe that our simple baseline approach of re-ranking using T5 strictly with the natural-language question against the paper title and abstract (Seq2seq T5 in Table~\ref{tab:main}) is more effective than the more involved approach employed by \texttt{covidex.t5}. When we apply the same date filtering to the official runs, we observe that the differences narrow. We also present \sys topping the Round~1 leaderboard in Table~\ref{tab:rnd1}. We observe again that our model excels at finding highly-relevant documents.}

To gain a better understanding of the impact of filtering the document collection to only articles published on or after January 1, 2020, we first compare the performance of \sys with and without the filter. Disregarding unjudged documents, it has an nDCG@10 of 0.668 ($-0.051$), P@5 of 0.777 ($-0.069$) and P@5 (F) of 0.589 ($-0.108$). All these differences are statistically significant. By far the largest reduction is on fully-relevant P@5, meaning that it can be more difficult to find highly relevant documents when considering the full document collection. We observed similar trends for BM25, with and without the 2020 filter. These trends also align with observations we made from the judgments themselves; we find that only 16\% of judged documents from prior to 2020 were considered relevant (with only 5\% fully relevant). Meanwhile, 32\% of judged documents after 2020 were considered relevant (19\% fully relevant).

\section{Conclusion}

In this work, we present \sys, an adaptation of a neural ranking pipeline for COVID-19 scientific literature search. The approach is zero-shot and adapts to medical literature by filtering the training data, using a contextualized language model based trained on scientific text, and by filtering the document collection. \edit{The zero-shot setting is important because it suggests that the approach can be generally applied to similar problems---even when no training data are available (which can be expensive to collect).} Through experiments and analysis on TREC-COVID, we find that each component of our approach is beneficial, and it outperforms or is comparable to approaches that are trained or tuned on TREC-COVID judgments. These observations underscore the importance of properly considering the domain when building medical search engines. We hope that techniques like \sys can help overcome the global COVID-19 crisis.

\section*{Acknowledgements}

\edit{Experiments on T5 models were supported by TPU machines provided by Google. We thank the anonymous reviewers for their helpful feedback.}

\bibliographystyle{acl_natbib}
\bibliography{biblio}

\appendix

\clearpage

\section{Appendix}
\label{sec:supplemental}

\subsection{List of MedSyn exclusion terms}\label{sec:sup:medsyn-excl}

The following terms were excluded from MedSyn when filtering MS-MARCO to reduce false positive matches: gas, card, bing, died, map, fall, falls.

\subsection{Med-MARCO Examples}\label{sec:sup:filter-examples}

A random sample of 10 queries from the filtered Med-MARCO dataset:
\begin{itemize}
\item[-] 747605  what is fistula with salivary drainage
\item[-] 586569  what causes cirrhosis besides alcohol
\item[-] 925416  what would cause pain in left shoulder and right elbow
\item[-] 258186  how long does it take to show pregnancy test
\item[-] 845485  what is the salary of the governor of ms \textit{(false positive)}
\item[-] 1070398 why is hands swell when waking up
\item[-] 956309  when to worry about high temperature in adults
\item[-] 776140  what is nervous breakdown
\item[-] 750061  what is gastric ulcer
\item[-] 83842   cat's eye meaning \textit{(false positive)}
\end{itemize}

\subsection{TREC-COVID Run Descriptions}\label{sec:sup:rnd1}

\texttt{sab20.1.meta.docs}: Simple SMART vector run, Lnu docs and ltu queries. Separate inverted files for metadata and JSON docs. Final score = 1.5 * metadata score + JSON score. Full topics including narrative.

\vspace{1em}

\texttt{IRIT\_marked\_base}: We use a BERT-base (12 layers, 768 hidden size) fine-tuned on Ms Marco passage set. We use a full ranking strategy with two stages: in the first stage, we use Anserini Bm25+ RM3 to retrieve top-1000 candidate documents for each topic using an index on the title+abstract of the CORD-19 documents, then we use the fine-tuned BERT to re-rank this list.

\vspace{1em}

\texttt{CSIROmedNIR}: A neural index was built on the title, abstract fields of the COVID corpus alongside a traditional inverted index built on title, abstract and body text of the document. The neural index was built from the pooled classification token (1st token of the final BERT layer) using the covidbert-nli model (https://huggingface.co/gsarti/covidbert-nli) from the title, based off the sentence transformer (Reimers etal. Sentence-BERT, 2019). For the abstract,we took the Bag-of-Sentence approach where we averaged the individual sentence embeddings (sentence were segmented using segtok). All embeddings had a final dimension size of [1, 768]. We searched on the neural index using the query, narrative and question fields of the topics using the same embedding approach as with the document title embedding over the title and abstract neural index fields giving a total of 6 cosine similarity computations.  We combine BM25 scores from traditional search over a combination of query,narrative and question fields over all document facets (body, title, abstract), giving a total of 9 different query-facet combinations. We take the natural logarithm of the total BM25 score (to match the range of the cosine scores) which is then added the cosine scores: finalscore = log(sum of BM25 query-facet combs)+ cosine Scores Additionally, we filter the document by date. Documents created before December 31st 2019(before the first reported case) had their scores automatically set to zero.

\vspace{1em}

\texttt{mpiid5.run3}: We re-rank top-10000 documents returned by BM25 using the queries produced by Udel’s method. For there-ranking method, we use the ELECTRA-Base model fine-tuned on the MSMARCO passage dataset. Themodel is later fine-tuned on the TREC COVID round 1 full-text collection.

\vspace{1em}

\texttt{SparseDenseSciBert}: bm25+ann+scibert.0.33.teIn (ann-bm25 retrieval + scibert reranker): anserini TREC-COVID R2 Retrieval\#8 + med-marco ANN + med-marco SciBERT with COVD Mask-Lm fine-tuning

\vspace{1em}

\texttt{covidex.t5}: Reciprocal rank fusion of two runs:  Anserini r2.fusion1, reranked with medT5-3B; Anserini r2.fusion2, reranked with medT5-3B; Anserini fusion baselines for round 2: \url{https://github.com/castorini/anserini/blob/master/docs/expcovid.md} medT5-3B: a T5-3B reranker fine-tuned on MS MARCO then fine-tuned (again) on MS MARCOmedical subset.

\subsection{Model Training and Validation}\label{sec:sup:params}

\begin{tabular}{ll}
\toprule
Parameter & Value \\
\midrule
Train Hardware & QuadroRTX 8000 GPU \\
& CUDA version 10.1 \\
Train Dataset & Med-MARCO (this work) \\
Loss Fuction & Pairwise Cross-Entropy \\
& from \citet{Nogueira2019PassageRW} \\
Max. Query length & 60 \\
Max. Document Length & 2000 \\
Base Model & scibert-scivocab-uncased \\
BERT Learning Rate & $2\times10^{-5}$ \\
Final Layer Learning Rate & $1\times10^{-3}$ \\
Optimizer & Adam \\
Warm-up & None \\
Batch Size & 16 \\
Grad. Accumulation Size & 2 \\
Samples Validation & 512 \\
Patience & 20 \\
Validation Dataset & 200 from MS-MARCO train \\
Validation Metric & MRR@10 \\
Validation Re-rank & BM25 top 20 \\
Train and Validation Index & Lucene (via Anserini) \\
Index Stemming & Porter \\
BM25 Parameters & k1=0.9, b=0.4 (defaults) \\
\bottomrule
\end{tabular}

\end{document}